%% file: root.tex
\documentclass[letterpaper, 10 pt, conference]{ieeeconf}  

\IEEEoverridecommandlockouts                              

\usepackage{amsmath}
\newtheorem{asu}{Assumption}
\usepackage{stfloats}
\usepackage{subcaption}
\usepackage{bbold}
\usepackage{graphicx}
\usepackage{xcolor}
\usepackage{float} 
\usepackage{makecell}
\usepackage{adjustbox}
\usepackage{amsfonts} 
\usepackage{epsfig} 
\usepackage{mathptmx} 
\usepackage{times} 
\usepackage{amssymb}  
\let\labelindent\relax
\usepackage{enumitem}
\usepackage{multirow}
\usepackage{stmaryrd}
\usepackage{textcomp}
\usepackage{arydshln}
\usepackage[hidelinks]{hyperref}
\title{\LARGE \bf
Volumetric Semantically Consistent 3D Panoptic Mapping 
}

\begin{document}
%


\author{
    Yang Miao\textsuperscript{1}, 
    Iro Armeni\textsuperscript{1, 2}, Marc Pollefeys\textsuperscript{1,3}, and Daniel Barath\textsuperscript{1} \\
    \textsuperscript{1}ETH Zurich,
    \textsuperscript{2}Stanford University, 
    \textsuperscript{3}Microsoft
    \vspace{-6mm}
}
\twocolumn[{%
\renewcommand\twocolumn[1][]{#1}%
\maketitle
\begin{center}
    \centering
    \begin{minipage}{.22\textwidth}
        \centering
        \includegraphics[width=\linewidth]{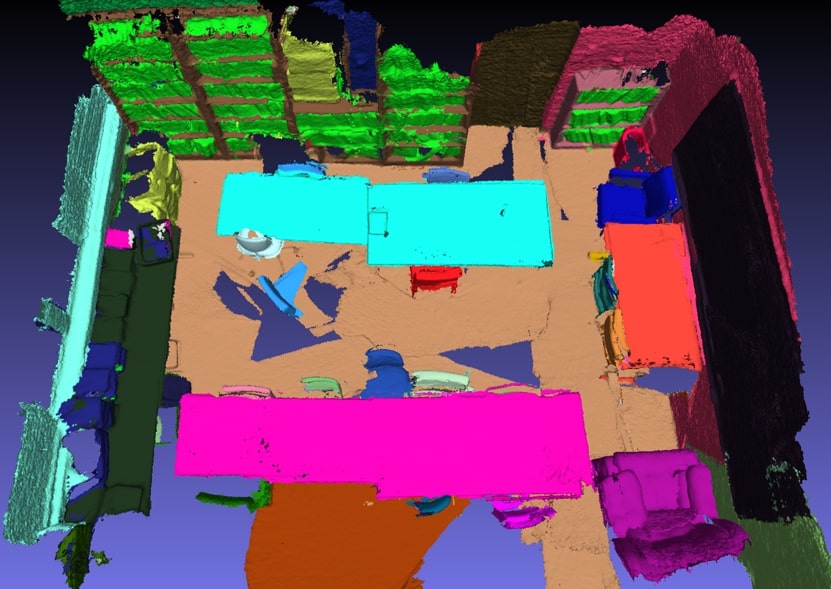} \\
        \vspace{1pt} Ours  
    \end{minipage}
    \hfill 
    \begin{minipage}{.22\textwidth}
        \centering
        \includegraphics[width=\linewidth]{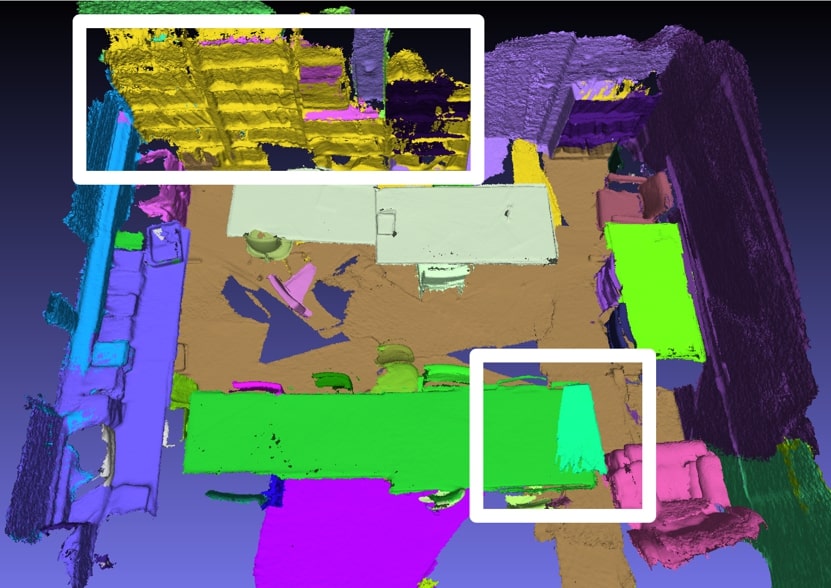} \\
        \vspace{1pt} Voxblox++~\cite{Semantic:voxbloxplusplus}
    \end{minipage}
    \hfill 
    \begin{minipage}{.22\textwidth}
        \centering
        \includegraphics[width=\linewidth]{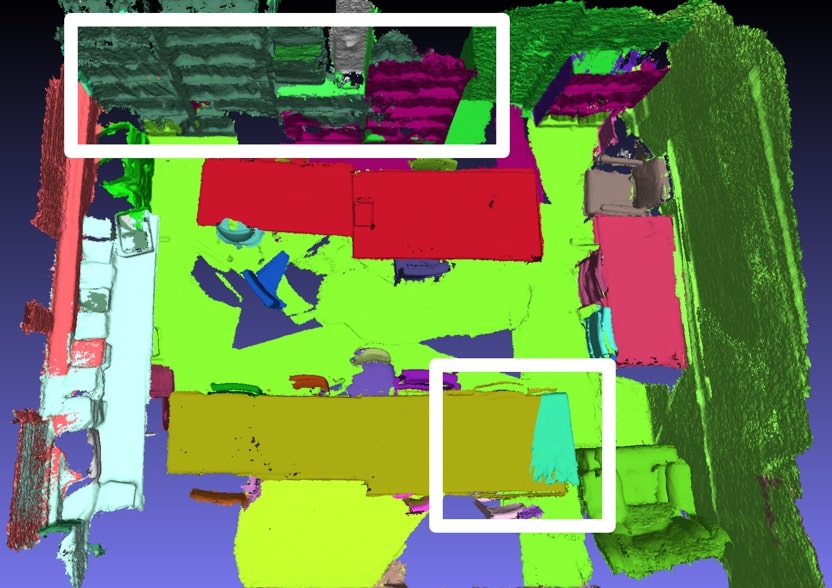} \\
        \vspace{1pt} Han et al.~\cite{ModifiedVoxblox}
    \end{minipage}
    \hfill 
    \begin{minipage}{.22\textwidth}
        \centering
        \includegraphics[width=\linewidth]{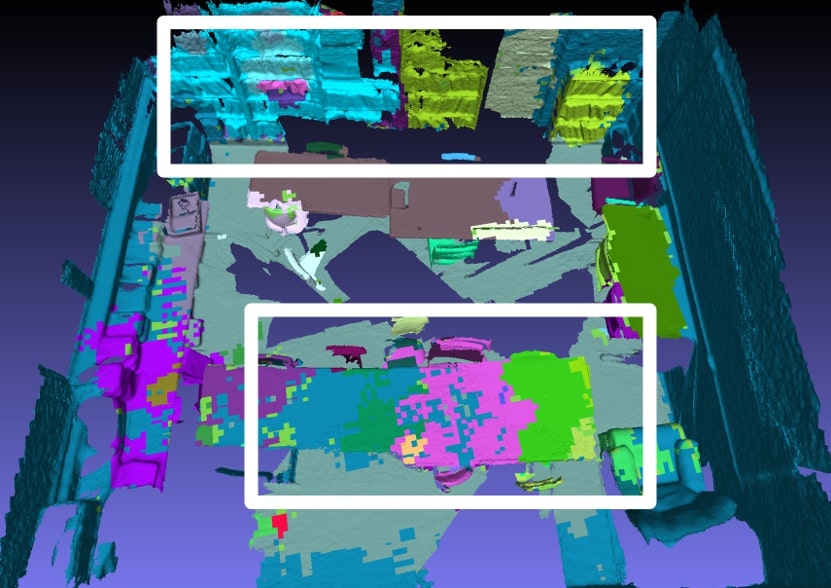} \\
        \vspace{1pt} INS-CONV~\cite{INS_CONV}
    \end{minipage}
    \captionof{figure}{
        3D panoptic segmentation comparison: the proposed method versus Voxblox++~\cite{Semantic:voxbloxplusplus}, Han et al.~\cite{Han20CVPR}, and INS-CONV~\cite{INS_CONV}. 
        In this example, all competitors under-segment the bookshelf (top rectangle) and over-segment the table (bottom).
        The proposed technique sets a new benchmark in 2D-to-3D semantic-instance segmentation by ensuring semantic consistency across super-points (a set of adjacent 3D voxels) through graph-based semantic optimization and instance refinement.
    }
    \label{fig:teaser}
\end{center}%
}]

\thispagestyle{empty}
\pagestyle{empty}


\begin{abstract}
We introduce an online 2D-to-3D semantic instance mapping algorithm aimed at generating comprehensive, accurate, and efficient semantic 3D maps suitable for autonomous agents in unstructured environments. 
The proposed approach is based on a Voxel-TSDF representation used in recent algorithms.
It introduces novel ways of integrating semantic prediction confidence during mapping, producing semantic and instance-consistent 3D regions.
Further improvements are achieved by graph optimization-based semantic labeling and instance refinement.
The proposed method achieves accuracy superior to the state of the art on public large-scale datasets, improving on a number of widely used metrics. 
We also highlight a downfall in the evaluation of recent studies: using the ground truth trajectory as input instead of a SLAM-estimated one substantially affects the accuracy, creating a large gap between the reported results and the actual performance on real-world data. 
The source code will be made public.
\end{abstract}

\section{INTRODUCTION}

\input{introduction.tex}
\section{RELATED WORK}

\input{related_work.tex}

\section{SEMANTIC-INSTANCE MAPPING}

\input{method.tex}

\section{EXPERIMENTS}

\input{experiments.tex}

\section{CONCLUSIONS}

In this paper, we introduce a novel 2D-to-3D semantic instance segmentation algorithm. 
Drawing upon a Voxel-TSDF representation, our method combines panoptic prediction confidence, semantically consistent super-points, and graph-optimized semantic labeling and instance refinement. 
These components allow the proposed algorithm to outperform current state-of-the-art techniques on public and large-scale datasets. 
We achieve this superior accuracy while maintaining real-time processing speeds comparable to less accurate alternatives. 
Although the proposed approach was inspired by Voxblox++, our results surpass its accuracy, on average, by a significant 16.1 mAP points, underscoring the impact of our improvements. 
Furthermore, we draw attention to the inherent pitfalls of the conventional practice of relying on ground truth camera poses as input as it introduces considerable disparities between reported outcomes and real-world performance.
We will release the source code.

\addtolength{\textheight}{-12cm}   







\bibliographystyle{./bib/IEEEtran} 
\bibliography{./bib/IEEEabrv,./bib/IEEEexample}

\end{document}

%% file: introduction.tex
Autonomous agents operating in unstructured environments require a comprehensive understanding of their surroundings to make informed decisions. 
In addition to perceiving the geometry of the environment, this includes understanding semantic categories and recognizing individual instances of objects within the scene.  
Robust perception of the surrounding complex space enables the agent to discover, segment, track, and reconstruct objects at the instance level to guide their subsequent actions. 
However, real-world scenarios exhibit considerable variability in object appearance, shape, and location, posing a challenge to robotic perception.

3D semantic-instance mapping is the problem of simultaneously detecting and segmenting objects with their semantic and instance labels to generate a 3D semantic map of the surrounding environment. There are two popular approaches to this problem: 
\textit{3D-to-3D}, which analyzes input 3D point clouds -- in this scenario the environment is previously reconstructed; and \textit{2D-to-3D}, which analyzes a set of 2D images, further projecting and aggregating predictions on the 3D map -- here, reconstruction and semantic mapping is done synchronously.
In practice, 3D-to-3D approaches often suffer from a lack of large-scale training data and speed issues, i.e., 3D data is more complex than 2D, which makes processing and analyzing 3D point clouds computationally intensive.
2D-to-3D algorithms leverage state-of-the-art 2D semantic-instance segmentation algorithms hence resolving some of the issues with 3D-to-3D approaches. However, having only 2D observations of the 3D instances, such methods often produce a noisy segmentation of the environment due to the limited perspective of 2D images and the noise in semantic detections per frame and in the camera calibration.

In this paper, we address the problem of noisy semantic detections by implementing semantic consistency and further regularization on semantics and instances. Specifically, we propose a 2D-to-3D semantic and instance mapping algorithm that achieves accuracy superior to state-of-the-art 2D-to-3D semantic instance segmentation methods on public large-scale datasets. Our approach is inspired by Voxblox++ \cite{Semantic:voxbloxplusplus}, which we improve upon with the following contributions:
\begin{enumerate}
    \item A new way of employing 2D semantic prediction confidence in the mapping process (Section~\ref{section:2d_segmentation}).
    \item Developing a novel method to segment semantic-instance consistent surface regions, i.e., super-points (Section~\ref{section:super_point}).
    \item A new graph optimization-based semantic labeling and instance refinement algorithm (Sections~\ref{section:segment_graph} \& \ref{section:instance_refinement}).
\end{enumerate}

%% file: related_work.tex
The task of 3D semantic instance mapping has a number of real-world applications in computer vision and robotics, allowing autonomous agents to understand their surroundings.
Methods used for this task can be divided into two main categories based on their input.

\textbf{3D-to-3D approaches} primarily analyze input 3D point clouds to discern instances directly in 3D.
Bottom-up approaches~\cite{Wang19CVPR, Elich19GCPR, Lahoud19ICCV, miao2021tianjiport} form object instances by grouping points based on their local affinities. 
In contrast, top-down approaches~\cite{Hou19CVPR, Yang19NIPS} initiate the process by identifying object bounding boxes and subsequently segmenting them into foreground and background regions. 
Another successful methodology is the voting-based approaches~\cite{Chen21ICCV, Engelmann20CVPR, Han20CVPR, Jiang20CVPR, Vu22CVPR}, where points nominate or "vote" for object centers, followed by a clustering process of the center votes. 
Recent Transformer-based strategies~\cite{schult2022mask3d, sun2022superpoint} represent objects using learned instance queries, correlating them with per-point features to create instance masks.

While these approaches can provide highly accurate results since they have access to the entire geometry of the scene and the objects, they often require a significant amount of computational resources and are challenging to scale to large scenes. In addition, they require a 3D map to be previously reconstructed, hence resulting in asynchronous methods.

\textbf{2D-to-3D approaches} take 2D images as input and progressively reconstruct the scene in a semantic and instance-aware manner. 
This approach has received increasing attention, especially in the robotics field, due to its capability to be real-time and incremental -- factors that are indispensable for applications such as navigation and path planning.

Among the representative techniques are Voxblox++ \cite{Semantic:voxbloxplusplus} and PanopticFusion \cite{PanopticFusion}.
Voxblox++ \cite{Semantic:voxbloxplusplus} employs both semantic instance and geometric segmentation, converting 2D frames into instance-labeled 3D segments. The process involves the incremental registration of geometric segments detected in depth frames onto global 3D counterparts, subsequently integrating semantic and instance labels within these global segments.
Our method is inspired by \cite{Semantic:voxbloxplusplus}, but it offers substantial enhancements, particularly in terms of semantic accuracy.
In \cite{PanopticFusion}, panoptic segmentation \cite{panoptic_segmentation} is applied to extract semantic instances for ``things'' and semantic labels for collective objects (``stuff"), which help improve the quality of semantic-instance segmentation.
SceneGraphFusion \cite{SceneGraphFusion} builds a scene graph with geometric segments and then applies a graph neural network to predict semantic and instance labels. However, SceneGraphFusion utilizes depth images synthetically rendered from a reconstructed 3D mesh, hence using ground truth data. 
The recent INS-CONV~\cite{INS_CONV} combines the advantages of 2D-3D and 3D-3D procedures. 
While it builds the 3D model of the environment incrementally from RGB-D frames, it performs segmentation directly in the 3D space. 
This, however, comes at the cost of requiring 3D ground truth annotations for training.

\begin{figure*}[!ht] 
\includegraphics[width=\textwidth]{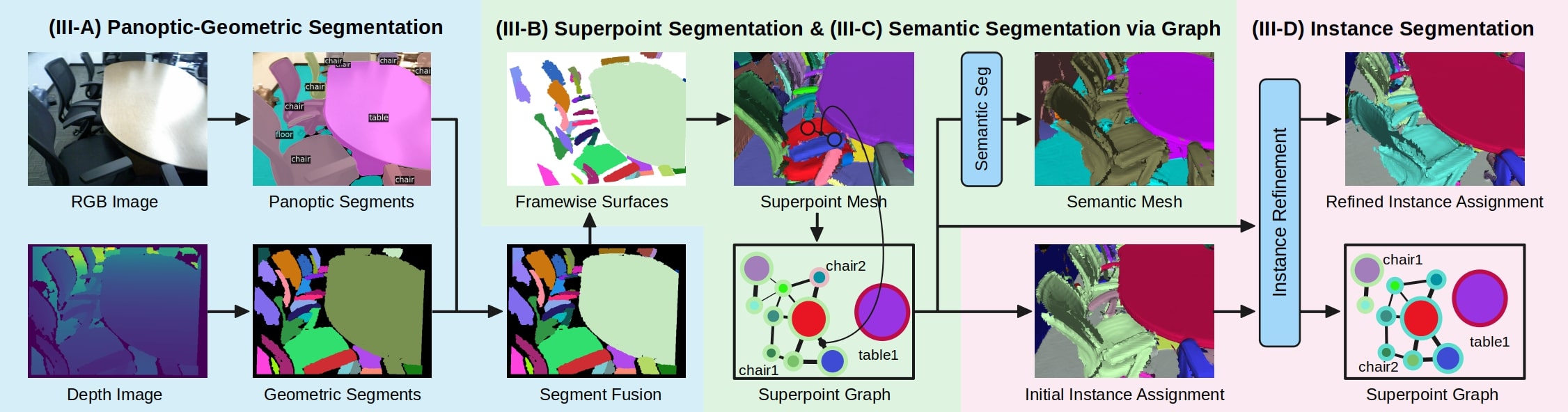}
\caption{
In \ref{section:2d_segmentation}, the proposed method gets an RGB-D sequence and runs panoptic and geometric segmentation that are then fused to extract 3D surfaces with panoptic labels. 
In \ref{section:super_point}, it incrementally lifts frame-wise surfaces into a global coordinate system to create semantically consistent superpoints.
In \ref{section:segment_graph}, superpoint graph is constructed and updated. Semantic segmentation is performed using the graph.
In \ref{section:instance_refinement}, the initial instance segmentation is refined by graph optimization.
}
\label{fig:pipeline}
\vspace{-10pt}
\end{figure*}

Noisy 2D predictions stand as a recurrent source of errors in 2D-to-3D techniques. Both geometric \cite{GeometricSeg} and panoptic segmentations \cite{mask_rcnn,panoptic_segmentation} occasionally yield over- or under-segmented scenes, leading directly to inaccuracies in the 3D segmentation.
To address geometric over-segmentation, Voxblox++ \cite{Semantic:voxbloxplusplus} adopts a strategy of iteratively merging proximate 3D segments. However, this method falls short in considering semantic consistency, resulting in incorrect mergers of disparate segments.
To rectify this issue, both \cite{PanopticFusion} and \cite{VoxbloxDiffusion} employ a computationally expensive voxel-level map regularization. 
Han et al. \cite{ModifiedVoxblox} make strides in the direction of leveraging spatial context, opting for regularization at the segment level. This, however, can only deal with 3D instance over-segmentation.
Distinctively, our approach incorporates super-point-level regularization and refinement. This not only ensures enhanced semantic-instance segmentation but also retains a small computational footprint.

%


%% file: method.tex
The proposed incremental 2D-to-3D instance and semantic mapping approach, illustrated in Fig.\ref{fig:pipeline}, processes a sequence of RGB-D frames alongside poses estimated by a SLAM pipeline \cite{campos2021orb}. 
The procedure encompasses a six-step process for each frame, delineated as follows:
(i) 2D panoptic-geometric segmentation, detailed in Sec.~\ref{section:2d_segmentation};
(ii) 3D super-point segmentation, described in Sec.~\ref{section:super_point};
(iii) super-point graph construction and update, elaborated in Sec.~\ref{section:segment_graph}.
Subsequent steps are executed at the end of the process:
(iv) semantic regularization, as discussed in Sec.~\ref{section:segment_graph};
(v) instance refinement, outlined in Sec.~\ref{section:instance_refinement};
and (vi) mesh generation, following the methodology presented in \cite{Semantic:voxbloxplusplus}.

\noindent
\textit{Preliminaries.}
Here, we define crucial concepts and notation utilized throughout this manuscript:
\begin{itemize}
    \item Panoptic segment $o$: A semantic-instance mask segmented from an RGB image, depicted in Fig.\ref{fig:pipeline} III-A.
    \item Geometric segment: A convex 3D patch segmented from a depth image, illustrated in Fig.\ref{fig:pipeline} III-A.
    \item Surface $s$: 3D points derived from a panoptic-geometric segment in an RGB-D frame, as shown in Fig.\ref{fig:pipeline} III-B.
    \item Super-point $S$: A collection of 3D voxels that represents a whole or part of object instance, labeled as $L_S$, and constructed from frame-wise surfaces $s$ (Fig.\ref{fig:pipeline} III-B).
    \item Super-point graph $\mathcal{G} = (\mathcal{V}, \mathcal{E})$: A graph comprising super-points ${S}$ as nodes and edges that model semantic and spatial relationships, as shown in Fig.~\ref{fig:pipeline} III-C, III-D.
\end{itemize}

\subsection{2D Panoptic-Geometric Segmentation} \label{section:2d_segmentation}
Panoptic-geometric segmentation, as depicted in Fig.~\ref{fig:pipeline} III-A, is applied to each RGB-D frame $F$. This process extracts convex object surfaces based on the assumption from \cite{GeometricSeg}:
\begin{asu}\label{convex_assumption}
    The surface of each real-world object is composed of single or multiple neighboring convex surfaces that share the same semantic-instance label.
\end{asu}

Initially, panoptic segmentation \cite{mask2former} is performed on the RGB image to produce a set of semantic instance masks. 
Concurrently, geometric segmentation \cite{GeometricSeg} is applied to the depth image to extract a set of convex segments. 
The intersections between the panoptic and geometric segmentation masks are utilized to generate a set of surface masks $\mathcal{F}_s = {s_i}$ for each frame $F$. 
Consequently, each surface mask $s_i$ is uniquely associated with a panoptic semantic instance $o_i$ and a geometric segment on a per-frame basis, thus mitigating the risk of geometric and semantic under-segmentation.

Each surface $s_i$ is attributed with an instance label $o_i$, a semantic category $C(o_{i})$, and a panoptic confidence score:
\begin{equation}\label{eq:super_point_inst_confidence}
P_{o_i} = \left\{
\begin{array}{rcl}
\text{Score}(o_i) \in (0,1)       &      & {C(o_i) \in C_\text{Th}}, \\
0.5     &      & {C(o_i) \in C_\text{St}}, \\
\end{array} \right.
\end{equation}
where $C_\text{Th}$ and $C_\text{St}$ denote the sets of "thing" and "stuff" categories, respectively, as defined in \cite{mask2former}. $\text{Score}(o_i)$ represents the confidence score for the instance $o_i$ from \cite{mask2former}. Utilizing the camera pose and the depth image, the point cloud $\text{Pcl}(s_i)$ corresponding to $s_i$ is projected onto a unified Voxel-TSDF structure, which is incrementally constructed.

\subsection{Semantically Consistent Super-point Segmentation}
\label{section:super_point}

In the context of volumetric mapping, surfaces are represented by voxels. 
Consequently, Assumption \ref{convex_assumption} extends from individual surfaces to encompass super-points:
\begin{asu}\label{super_point_assumption}
    Each real-world object comprises one or more spatially contiguous super-points, having the same semantic-instance label.
\end{asu}

\begin{figure}[t] 
\begin{center}
\includegraphics[width=0.8\columnwidth]{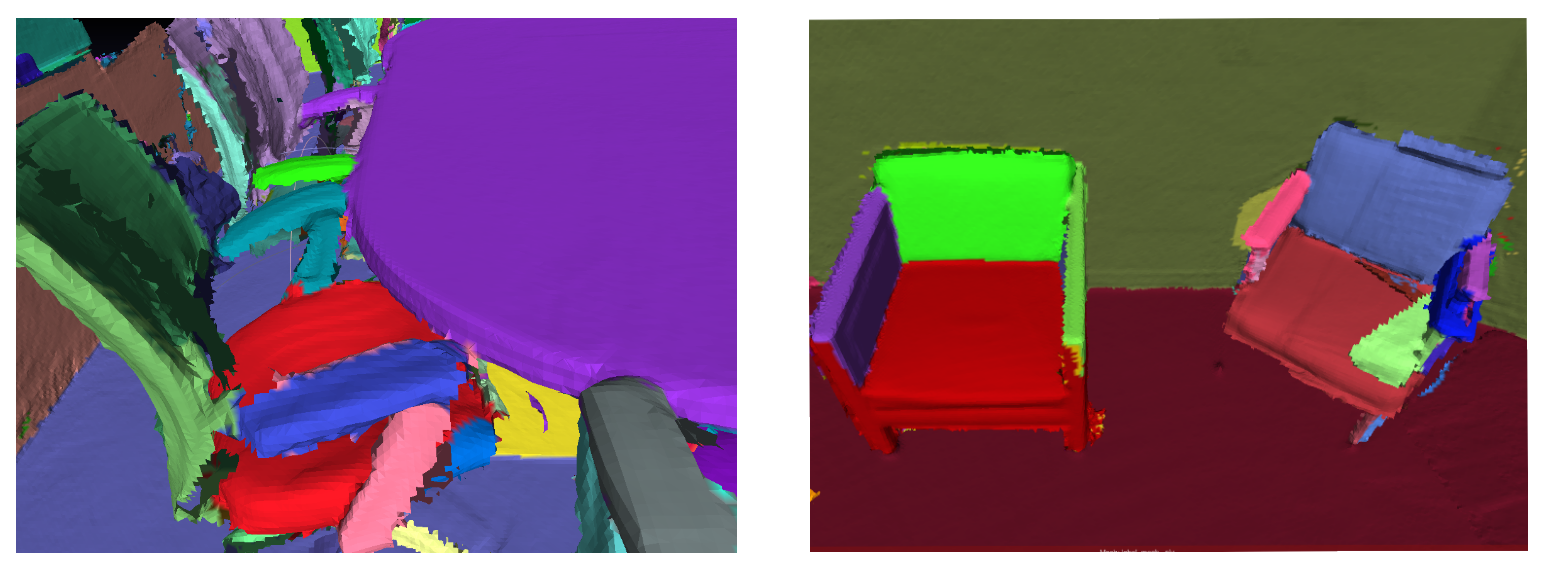}
    \caption{3D segments colored by super-point labels, depicting objects (e.g., chair, wall) composed of single or multiple super-points. Each super-point is unique to an object.}\label{fig:superpoint_seg}
\end{center}
\vspace{-15pt}
\end{figure}

Given Assumptions \ref{convex_assumption} and \ref{super_point_assumption}, we interpret each surface $s_i \in \mathcal{F}_s$ from Sec. \ref{section:2d_segmentation} as a partial observation of a super-point $S$ within a global 3D coordinate system. 
Consequently, super-points are registered or modified with frame-wise surface observations $s_i \in \mathcal{F}_s$ throughout the sequence of frames $\mathcal{F}$. 
The correlation between a surface $s_i$ and a super-point is established by computing the overlap $\Pi_s(L_S, s_i)$, which is the count of overlapping voxels between $s_i$ and super-point $L_S$ in the Voxel-TSDF map.
For every point $p_i$ in $\text{Pcl}(s_i)$, the corresponding voxel $V(p_i)$ is queried for its super-point label $L_S$, incrementing the overlap measure as $\Pi_s(L_S(V(p_i)), s_i) := \Pi_s(L_S(V(p_i)), s_i) + 1$. 
A surface $s_i$ is associated with the super-point labeled $L_S$ if the overlap surpasses a predefined threshold. 
In the absence of sufficient overlap, $s_i$ receives a new label $L_{S_{\text{new}}}$. 
Fig. \ref{fig:superpoint_seg} shows that the super-points segmentation meets Assumption \ref{super_point_assumption}.

To validate Assumption \ref{super_point_assumption}, we track $\Pi_V(L_S, V_i)$, the occurrence count of each candidate super-point with label $L_S$ in voxel $V_i$. The super-point label $L_S^*$ for voxel $V_i$ is determined as follows:
\begin{equation}\label{eq:voxel_label}
L_S^*(V_i) = \arg \max_{L_S}\Pi_V(L_S, V_i),
\end{equation}
indicating that a voxel is allocated to the super-point receiving the majority of votes. This equation is applied when integrating a new surface $s_i$ into the global map and querying the super-point label $L_S$ in voxel $V(p_i)$.

To minimize over-segmentation, super-points $S_1$ and $S_2$ are merged if they are spatially connected and semantically consistent, fulfilling these criteria by being:
\begin{enumerate}
\item \textit{Spatially connected}: $M(L_{S_1}, L_{S_2}) > \theta_\text{merge}$;
\item \textit{Semantically consistent}: Semantic categories ${C}(L_{S_1})$ and ${C}(L_{S_2})$ either match or at least one matches a background category $C_0$. Function ${C}$ maps super-points to their semantic labels, as defined in Sec.~\ref{section:segment_graph}.
\end{enumerate}
The spatial connection measure $M(L_{S_1}, L_{S_2})$ quantifies the 3D overlap between super-points $S_1$ and $S_2$, aggregated over frames ${\mathcal{F}}$, as follows:
\begin{equation}
    \small
    M(L_{S_1}, L_{S_2}) = \sum_{\{\mathcal{F}\}} \sum_{s_i \in \mathcal{F}_s} \llbracket S_1 \in L_N(s_i) \wedge S_2 \in L_N(s_i) \rrbracket,
\end{equation}
where $L_{N}(s_i)$ denotes the set of super-points significantly overlapping with $s_i$, and $ \llbracket . \rrbracket$ is the Iverson bracket which equals to $1$ if the condition inside holds and $0$ otherwise.
\begin{figure}[t] 
\begin{center}
\includegraphics[width=0.8\columnwidth]{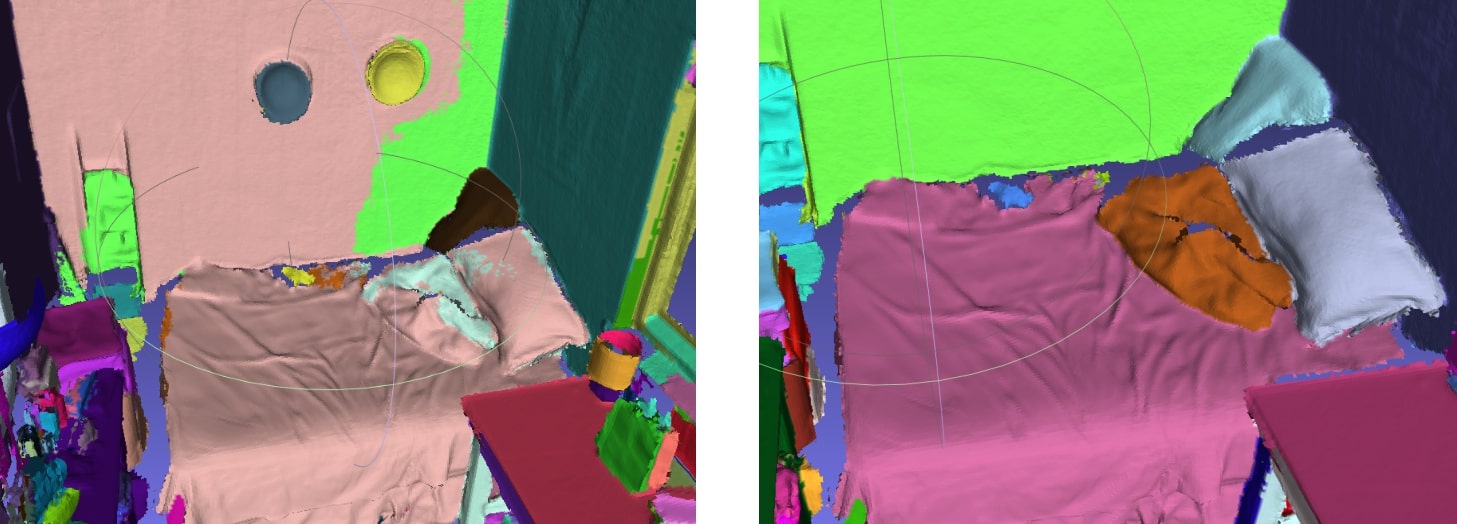}
    \caption{Super-point segmentation without (left; as in \cite{Semantic:voxbloxplusplus}) and with (right) semantic consistency checks as proposed in Sec.~\ref{section:super_point}. 
    Without semantic consistency, noisy geometric segmentations \cite{GeometricSeg} result in merging the super-points of "wall" and "bed".
    By distinguishing semantic labels as proposed, such incorrect mergers happen less often.} \label{fig:ablation_super-point}
\end{center}
\vspace{-15pt}
\end{figure}  

Integrating semantic consistency into super-point segmentation (example in Fig.~\ref{fig:ablation_super-point}), ensures adherence to Assumption \ref{super_point_assumption}, resulting in enhanced segmentation accuracy. This approach improves upon methods such as \cite{Semantic:voxbloxplusplus, ModifiedVoxblox}, which primarily consider spatial overlap. The outcome of this step is a refined set of super-points within the RGB-D frame $\mathcal{F}$.


\subsection{Semantic Segmentation via Super-Point Graph}
\label{section:segment_graph}

The subsequent step of estimating super-points for each frame involves translating them into the global coordinate system and associating them with global object instances.

\begin{figure}[t] 
\begin{center}
\includegraphics[width=0.8\columnwidth]{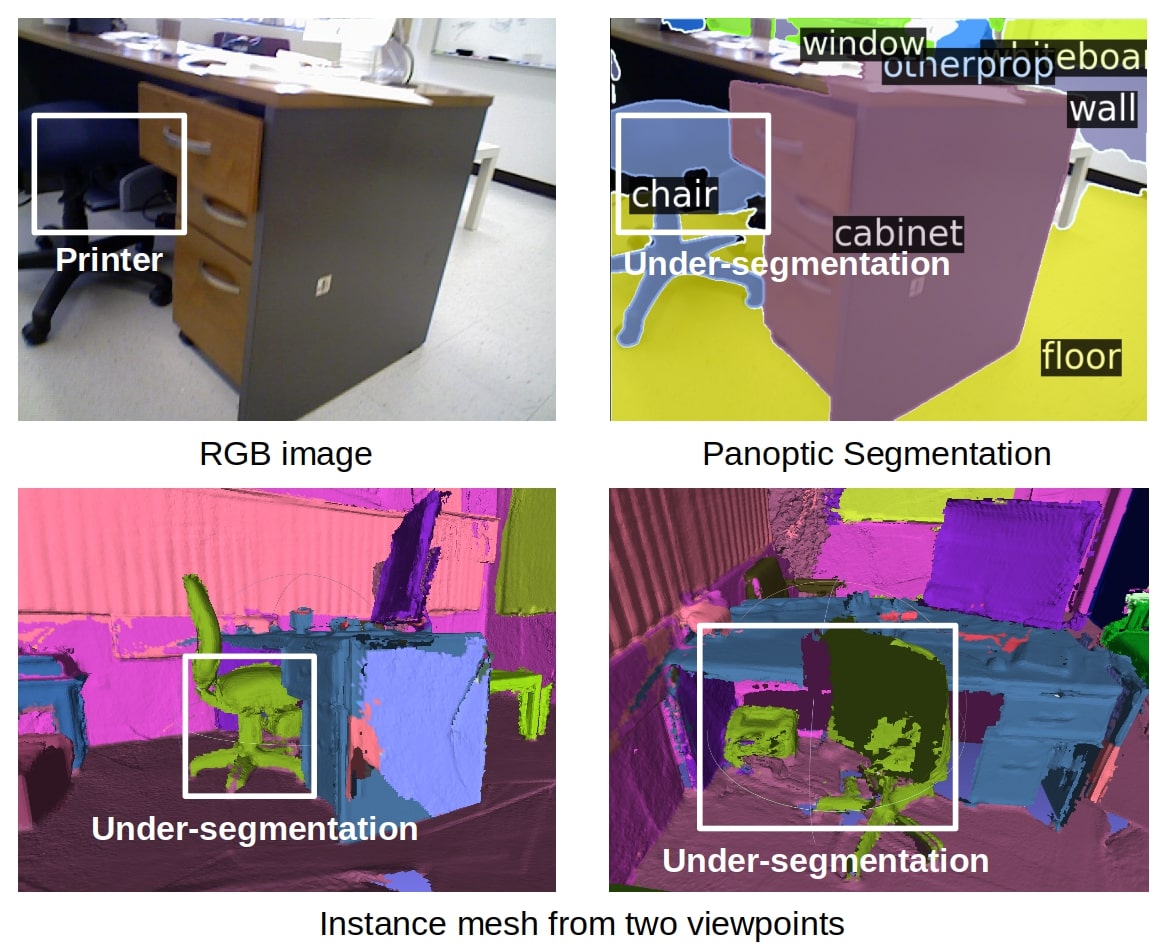}
    \caption{Under-segmentation of the chair and the printer in the corner using Voxblox++ \cite{Semantic:voxbloxplusplus}, due to inaccuracies in panoptic segmentation. Instances are shown by color.}\label{fig:underseg}
\end{center}
\vspace{-15pt}
\end{figure}

The initial semantic and instance labeling of super-points leverages the frame counting method used in \cite{ModifiedVoxblox, Semantic:voxbloxplusplus} with each frame of the sequence voting for certain semantic and instance labels for super-points observed in that frame.
This technique utilizes a 3D overlap-based data association to map frame-wise 2D instances ${o}$ to 3D global instance labels ${O}$. 
It forms frame-wise associations of super-points and global instances, assigning super-points to the global label with the highest frame count overlap.
We refer readers to \cite{ModifiedVoxblox} for an in-depth understanding. 
This approach, however, exhibits several limitations:
(i) By simplistically assigning super-points to the global instance labels with maximum frame counts, it neglects that the noise of panoptic-geometric segmentation could be different across frames due to viewpoint differences;
(ii) The method solely relies on individual observations of each super-point without considering the semantic-instance correlations among super-points. 
Consequently, noisy 2D panoptic-geometric predictions with high frame counts may erroneously influence label assignments, leading to incorrect semantic-instance labeling of super-points, such as over/under-segmentation of instances. 
An example of this issue is depicted in Fig.~\ref{fig:underseg}.

To address these challenges, we employ a graph-based representation to collectively process frame-wise semantic-instance data and mitigate the impact of noisy 2D predictions by assigning lower panoptic-spatial confidence to such frames. 
Subsequently, graph-based optimization refines the semantic-instance labels by leveraging the aggregated global panoptic-geometric information within the super-point graph. 
Specifically, a graph $\mathcal{G} = (\mathcal{V}, \mathcal{E})$ is formulated with super-points serving as vertices ($S \in \mathcal{V}$). 
Each vertex maintains a semantic score $P_{\nu}^C(L_{S})$, indicating the likelihood of vertex with label $L_{S}$ being associated with semantic category $C$. 
To represent panoptic-spatial relationships between super-points, an edge $\epsilon^C = (L_{S_1}, L_{S_2}) \in \mathcal{E}$ connects two vertices if the panoptic-spatial confidence $P_{\epsilon}^C(L_{S_1}, L_{S_2}) > 0$. 
This confidence score, $P_{\epsilon}^C(L_{S_1}, L_{S_2})$, quantifies the probability of two super-points $L_{S_1}$ and $L_{S_2}$ sharing the instance label within semantic class $C$ and are spatially contiguous.

\textit{Confidence calculation.}
Here, we detail the computation of confidences $P_{\nu}^C$ and $P_{\epsilon}^{C}$. 
For each frame, rays projected from the camera center through the 2D panoptic masks $o_i$ intersect with the 3D super-point map. Voxels intersected by these rays are grouped into sets $\{V_{o_i}^{L_S}\}$ based on their super-point label $L_S$, as determined by Eq. \ref{eq:voxel_label}. Following this, $P_{\epsilon}^C$ and $P_{\nu}$ are updated with information from the current frame.

For the semantic score $P_{\nu}^C$, the confidence assigned to a class $C(o_i)$ for a super-point $S$ is calculated by aggregating the intersected voxels of $S$ with the rays from mask $o_i$:
\begin{equation} \label{eq:semantic_score}
P_{\nu}^C(L_S) = P_{\nu}^C(L_S) + P_{o_i} \cdot N^S_{o_i},
\end{equation}
where $N^S_{o_i}$ is the voxel count in set $V_{o_i}^{L_S}$. 
This implies that the more voxels from super-point $L_S$ intersect with rays from the panoptic mask $o_i$, the higher the likelihood that super-point $L_S$ pertains to the semantic category of $o_i$.

Panoptic-spatial confidence $P_{\epsilon}^{C}$ is updated for each pair of super-points $L_{S_1}$ and $L_{S_2}$ whose voxels intersect with rays from the same panoptic mask $o_i$. This indicates the likelihood that the two super-points are part of the same object $o_i$:
\begin{equation*} \label{eq:pairwise_score}
\small
    P_{\epsilon}^{C(o_i)}(L_{S_1}, L_{S_2}) := P_{\epsilon}^{C(o_i)}(L_{S_1}, L_{S_2}) + P_{o_i} \cdot P_{s} \cdot \min(N^{S_1}_{o_i}, N^{S_2}_{o_i}),
\end{equation*}
where $P_{o_i}$ is the panoptic confidence of $o_i$, as defined in Eq. \ref{eq:super_point_inst_confidence}, and $P_{s}$ is the spatial confidence based on the inverse distance of $L_{S_1}$ and $L_{S_2}$. 
This approach suggests that the greater the panoptic and spatial confidences, the more probable the two super-points observed within the frame belong to the instance $o_i$.

\textit{Graph optimization.}
Here, semantic segmentation based on graph optimization is discussed. 
The goal is to refine the mapping of super-points to semantic labels by constructing and minimizing an energy function $E(C)$, where $C$ represents the semantic labels assigned to super-points. This optimization employs the $\alpha-\beta$ swap algorithm \cite{GraphCut}.

The energy function incorporates both unary and binary potentials, formulated as:
\begin{equation} \label{eq:energy_fuction}
    \small
    E(C) = \sum_{L_S \in \mathcal{V}} \psi_u(C(L_S)) + \sum_{(L_{S_1}, L_{S_2}) \in \mathcal{E}} \psi_p(C(L_{S_1}), C(L_{S_2})).
\end{equation}
Unary potential $\psi_u(C(L_S))$ reflects the cost associated with assigning super-point $S$ to semantic category $C(L_S)$. It is inversely proportional to the probability of $C(L_S)$:
\begin{equation} \label{eq:unary_energy}
    \begin{split}
        & \psi_u(C(L_S)) = -\log \frac{ P_{\nu}^{C(L_{S})}(L_{S}) }{\sum_{C}  P_{\nu}^C(L_{S}) },
    \end{split}
\end{equation}
with $\sum_{C}$ indicating the sum over all semantic labels.

The binary potential, $\psi_p(C(L_{S_1}), C(L_{S_2}))$, represents the cost when two adjacent super-points are assigned different semantic labels. This potential is defined as:
\begin{equation} \label{eq:binary_energy}
    \small
    \begin{split}
        & \psi_p(C(L_{S_1}), C(L_{S_2})) = \mathbb{1}_{C(L_{S_1}) \neq C(L_{S_2})} K_{C} \exp{\left(-\frac{k(*)}{2\theta^2} \right)}, \\
        & k(L_{S_1}, L_{S_2}, C) = 
            \frac{{P_\nu}^{C(L_{S_1})}({L_{S_1}}) {P_\nu}^ {C(L_{S_2})}({L_{S_2}})}{\sum_{C} P_{\epsilon}^{C}(L_{S_1}, L_{S_2}) },
    \end{split} 
\end{equation}
where $k(L_{S_1}, L_{S_2}, C)$ evaluates the likelihood of $L_{S_1}$ and $L_{S_2}$ originating from different semantic categories, based on their panoptic-spatial scores $P_{\epsilon}^{C}(L_{S_1}, L_{S_2})$. A lower $P_{\epsilon}^{C}(L_{S_1}, L_{S_2})$ score implies a higher probability of differing semantic-instance labels, thereby reducing $k(L_{S_1}, L_{S_2}, C)$ and, consequently, the cost $\psi_p(C(L_{S_1}), C(L_{S_2}))$.

The energy function $E(C)$ is designed to be submodular and semi-metric, allowing for efficient minimization through the $\alpha-\beta$ swap algorithm $C^* = \arg \min_{C} E(C)$.
%
%
This optimization refines the semantic labels of super-points, enhancing the overall accuracy of the semantic segmentation.

\begin{table*}[!ht]
\color{black}
\centering
\caption{The mAP50 scores (higher is better) on 10 scenes of the SceneNN dataset~\cite{SceneNN} using the GT (top) and SLAM-estimated~\cite{campos2021orb} (bottom) camera poses. 
Results on GT are from \cite{VoxbloxDiffusion}. 
The SLAM part shows methods with public code.}
\vspace{-2mm}
\label{table:scenenn_10}
\begin{center}
\small
\resizebox{0.8\textwidth}{!}{
\begin{tabular}{c | l | c c c c c c c c c c | c}
\hline
&Method / Sequences                                & 011 & 016 & 030 & 061 & 078 & 086 & 096 & 206 & 223 & 255 & Average \\
\hline
\multirow{7}{*}{\rotatebox[origin=c]{90}{\scriptsize GT trajectory}} 
& Voxblox++ \cite{Semantic:voxbloxplusplus}   & \phantom{1}75.0& 48.2& 62.4& 66.7& \phantom{1}55.8&20.0& 34.6& 79.6& 43.8& \phantom{1}75.0& 56.1\\
&Han et al. \cite{ModifiedVoxblox}           & \phantom{1}65.8& 50.0& 66.6& 43.3& \textbf{100.0}& 56.9& 22.8& \textbf{92.1}& 46.7& \phantom{1}33.0& 57.7\\
&Wang et al. \cite{MultiviewFusion}          & \phantom{1}62.2& 43.0& 60.7& 36.3& \phantom{1}49.3& 45.8& 32.7& 46.0& 46.6& \phantom{1}56.4& 47.9\\
&Li et al. \cite{IncrementalBBox}            & \phantom{1}78.6& 25.0& 58.6& 46.6& \phantom{1}69.8& 47.2& 26.7& 78.0& 45.8& \phantom{1}75.0& 55.1\\
&Mascaro et al. \cite{VoxbloxDiffusion}      & \textbf{100.0}& \textbf{75.0}& 72.5& 50.0& \phantom{1}50.0& 50.0& 51.3& 74.1& 45.8& \textbf{100.0}& 66.9\\
&INS-CONV \cite{INS_CONV}                   & \textbf{100.0}& 62.0& 83.4& \textbf{69.8}& \phantom{1}93.7& 60.0& 57.6& 56.7& \textbf{78.6}& \textbf{100.0}& 76.2\\
&Ours                                        & \textbf{100.0}& 73.3& \textbf{91.7}& 62.4& \phantom{1}87.5& \textbf{61.7}& \textbf{66.7}& 83.3& 60.0& \textbf{100.0}& \textbf{78.7}\\
\hline
\multirow{4}{*}{\rotatebox[origin=c]{90}{\scriptsize SLAM}} 
& Voxblox++ \cite{Semantic:voxbloxplusplus}
& \phantom{1}61.5& 38.9& 50.0 & 58.4 & \phantom{1}44.3 & 16.4 & 27.6 & 48.7 & 40.7 & \phantom{1}33.6& 42.0\\
& Han et al \cite{ModifiedVoxblox}
& \phantom{1}53.4& 43.2& 50.0 & 37.6 & \phantom{1}75.6 & 48.2 & 13.4 & \textbf{57.8} & 44.1 & \phantom{1}24.7& 44.8\\
& INS-CONV \cite{INS_CONV}   
& \phantom{1}\textbf{75.0}& 46.7& 56.4& 57.1& \phantom{1}\textbf{83.7}& 22.4& \textbf{48.1}& 28.1& \textbf{65.2}& \phantom{1}50.0 & 53.3 \\
& Ours
& \phantom{1}\textbf{75.0}& \textbf{56.7}& \textbf{72.3}& \textbf{62.4}& \phantom{1}68.9& \textbf{55.6}& 33.2& 40.8& 60.0& \phantom{1}\textbf{63.3} & \textbf{58.8}\\
\hline
\end{tabular}}
\end{center}
\vspace{-10pt}
\end{table*}
\begin{figure}[t] 
\begin{center}
\includegraphics[width=0.8\columnwidth]{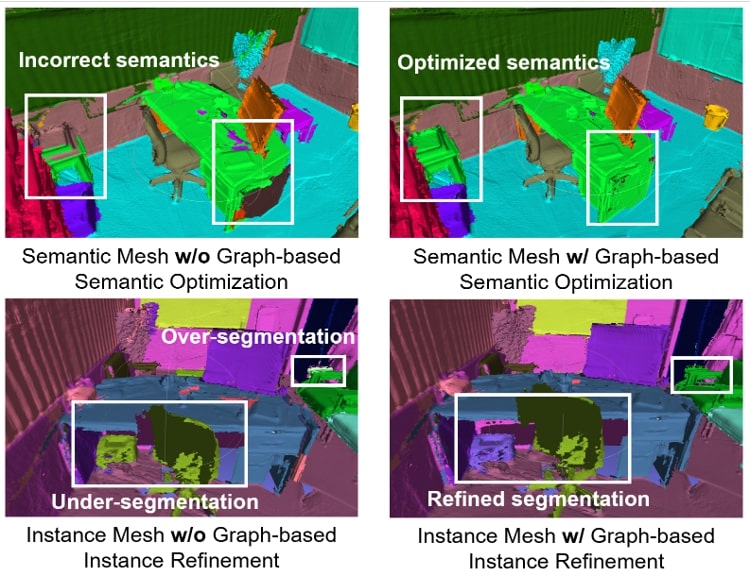}
    \caption{
    The generated mesh without (left) and with (right) the proposed graph-based semantic optimization (Sec.~\ref{section:segment_graph}) and instance label refinement (Sec.~\ref{section:segment_graph}).} \label{fig:ablation_graph_based_seg}
\end{center}
\vspace{-15pt}
\end{figure} 

\subsection{Instance Segmentation via Super-Point Graph} \label{section:instance_refinement}
This section delves into enhancing instance segmentation to address the challenges of under-segmentation and over-segmentation observed in existing methods like \cite{Semantic:voxbloxplusplus,ModifiedVoxblox}. 
Under-segmentation occurs when super-points from distinct instances are incorrectly merged under a single instance label. 
Over-segmentation results in a single instance being split across multiple labels due to errors in panoptic-geometric predictions or inaccuracies in data association from camera poses and depth measurements.

To address these issues, we now focus on refining instance labels. 
Instance label refinement is conducted for each semantic class, allowing for focused adjustment within super-point sets on specific semantic labels. 
Unlike semantic categories, which are finite, the range of instance labels is potentially infinite, expanding as new instances are detected during the mapping process. 
Therefore, the optimization strategies applicable to semantic labeling are unsuitable for instance segmentation, necessitating a different approach.

Our method employs rule-based algorithms for assigning instance labels to super-points. 
We start with a set of super-points $\mathcal{V}_C = \{ L_S \; | \; C(L_S) = C \}$ semantically linked to class $C$, along with their initial instance assignments $\mathcal{O}_C = { O_{L_S} }$, where ${L_S} \in \mathcal{V}_C$. The objective is to refine $\mathcal{O}_C$ by clustering super-points with strong interconnections (high panoptic-geometric scores $P{\epsilon}^C$) under the same instance label and segregating those with weak links.

An iterative process is employed, beginning with defining the instance confidence $P_O(O^C_i)$ for each unique instance label $O^C_i \in \mathcal{O}_C$ as the maximum edge confidence within the semantic class $C$:
\begin{equation} \label{instance_confidence}
P_O(O^C_i) = \max_{(L_{S_1}, L_{S_2}) \in \mathcal{E}_C(O^C_i)} P_{\epsilon}^{C}(L_{S_1}, L_{S_2}),
\end{equation}
where $\mathcal{E}_C(O^C_i)$ denotes the set of edges connecting super-points assigned to $O^C_i$. Super-points $L_S \in \mathcal{V}_C$ are then evaluated to determine if they should remain within their initial instance grouping $O^C_i$ based on the strength of their connections to other super-points within the same instance. Those with insufficiently strong connections (below a threshold $\theta^C_D P_O(O^C_i)$) are considered for reassignment.

For each super-point $L_S$ undergoing reevaluation, we explore its neighbors to identify potential reassignments to instances with strong connections, satisfying conditions $P^C_{L_S}(O_\text{neigh}) > \theta^C_O P_O(O_\text{neigh})$ and $P^C_{L_S}(O_\text{neigh}) > \theta^C_L P_L^C(L_S)$. 
Neighbors are defined based on spatial-panoptic connectivity with $P_{\epsilon}^C(L_S, L_\text{neigh}) > 0$. Super-points without suitable neighbors are allocated a new instance label $O_\text{new}$.

%% file: experiments.tex
We conduct experiments and compare the proposed method with multiple state-of-the-art frameworks~\cite{VoxbloxDiffusion,ModifiedVoxblox,Semantic:voxbloxplusplus,MultiviewFusion,IncrementalBBox,INS_CONV} on the SceneNN~\cite{SceneNN} and ScanNet v2 datasets~\cite{ScanNet}, containing RGB-D sequences, ground truth (GT) 3D meshes, camera poses, and semantic instance annotations. 
We use four metrics to compare accuracy, computed by thresholding the intersection over union (IoU) at thresholds $0.5$ and $0.75$: 
(i) mean average precision ($\text{mAP}$), 
(ii) total number of true positive predicted instances ($N_\text{TP}$), 
(iii) panoptic quality ($\text{PQ}$) \cite{panoptic_segmentation}, 
and the (iv) Intersection over Union between GT instance and segmented super-points ($\text{IoU}_{L_S}$), defined as 
%
    $\text{IoU}_{L_S} = \frac{1}{\left|O\right|} \sum_{o \in O} \sum_{L_S} \text{IoU}(L_S, o)$,
%
where $O$ represents the set of all GT instances. 
Metric $\text{IoU}_{L_S}$ measures the segmentation accuracy of super-points, and is only for methods performing super-point segmentation and having public source code~\cite{Semantic:voxbloxplusplus,ModifiedVoxblox}.
We use the \cite{mask2former} panoptic segmentation network for all methods. 

As per standard practice, we run methods using GT camera poses. 
Additionally, we run all approaches on poses estimated by ORB-SLAM3~\cite{campos2021orb} to demonstrate the significant performance gap between the practical setting (poses from SLAM) and the practice followed in recent papers (GT pose). 

For SLAM-based experiments, we limit our evaluations to \cite{Semantic:voxbloxplusplus,ModifiedVoxblox,INS_CONV}, as these are the sole methods with public code. 
Since \cite{INS_CONV} does not offer training code and their model has been trained on the ScanNet training set, we ensure consistency by fine-tuning or training all methods on ScanNet for the SLAM-based evaluations. Consequently, these experiments also serve as a valuable test of generalization.


\begin{table*}
\setlength\extrarowheight{0.5pt}
\centering
\caption{
Results on the entire validation set (20 scenes) of the SceneNN dataset \cite{SceneNN} using the GT (top) and SLAM-estimated~\cite{campos2021orb} camera poses as input.  
The reported accuracy metrics are the class-averaged $\text{mAP}$, class-summed $N_\text{TP}$, panoptic quality (\text{PQ}) and class-averaged super-point accuracy (IoU$_{L_S}$).
%
}
\vspace{-3mm}
\label{table:scenenn_74}
\begin{center}

\begin{subtable}{1.0\linewidth}\centering
\caption{Comparison of methods.}\label{table:scenenn_10_val}
\vspace{-1mm}
\small
\begin{tabular}{c | l | c c | c c | c c | c }
\hline
&Method  &  $\text{mAP}_{50} $  &  $\text{mAP}_{75} $ ($\uparrow$) & $N_{\text{TP}_{50}}$ & $N_{\text{TP}_{75}}$ ($\uparrow$) & $PQ_{50}$ & $PQ_{75}$ ($\uparrow$) & $\text{IoU}_{L_S}$ ($\uparrow$)\\ 
\hline
\multirow{4}{*}{\rotatebox[origin=c]{90}{\tiny GT trajectory}} 
&Voxblox++ \cite{Semantic:voxbloxplusplus}
& 33.0 & 24.6 & 253 & 191 & 21.4 & 15.7 & 64.2 \\
&Han et al. \cite{ModifiedVoxblox}
& 40.3 & 32.7 & 286 & 242 & 35.6 & 27.9 & 73.8\\
&INS-CONV \cite{INS_CONV}
& \textbf{51.9} & \textbf{38.2 }& \textbf{341} & \textbf{266} & 42.4 & \textbf{33.4} & -- \\
&Ours 
& 51.4 & 35.8 & 329 & 243 & \textbf{43.8} & 29.4 & \textbf{82.5}\\
\hline
\multirow{4}{*}{\rotatebox[origin=c]{90}{\tiny SLAM}} 
&Voxblox++ \cite{Semantic:voxbloxplusplus}
& 25.7& 17.6 & 188 & 169 & 16.2 & 13.0 & 56.8 \\
&Han et al. \cite{ModifiedVoxblox}
& 27.9& 16.8 & 204 & 165 & 15.4 & 11.9 & 61.3\\
&INS-CONV \cite{INS_CONV}
& 37.7 & 25.9 & 277 & 194 & 18.5 & 15.7 & -- \\
&Ours
& \textbf{41.4} & \textbf{30.7} & \textbf{304} & \textbf{216} & \textbf{23.7} & \textbf{17.4} & \textbf{74.3} \\
\hline
\end{tabular}%
\end{subtable}
\vspace{0.3pt}

\begin{subtable}{1.0\linewidth}\centering
\caption{Ablation study.}\label{table:scenenn_10_ablation}
\vspace{-1mm}
\small
\resizebox{0.8\textwidth}{!}{
\begin{tabular}{c | l | c c | c c | c c | c }
\hline
& Method  &  $\text{mAP}_{50} $  &  $\text{mAP}_{75} $ ($\uparrow$) & $N_{\text{TP}_{50}}$ & $N_{\text{TP}_{75}}$ ($\uparrow$) & $PQ_{50}$ & $PQ_{75}$ ($\uparrow$) & $\text{IoU}_{L_S}$ ($\uparrow$) \\ 
\hline
\multirow{8}{*}{\rotatebox[origin=c]{90}{\scriptsize GT trajectory}} 
& 1. Ours
& \textbf{51.4} & \textbf{35.8} & \textbf{329} & \textbf{243} & \textbf{43.8} & \textbf{29.4} & \textbf{82.5}\\
& 2. Ours  w/o \ref{section:super_point}
& 42.8 & 30.1 & 295 & 212 & 32.3 & 27.0 & 68.7 \\
& 3. Ours  w/o \ref{section:segment_graph}
& 48.6 & 34.1 & 311 & 231 & 36.2 & 28.2 & \textbf{82.5} \\
& 4. Ours  w/o \ref{section:instance_refinement}
& 47.9 & 33.7 & 307 & 227 & 35.7 & 27.8 & \textbf{82.5}\\
& 5. Ours  w/o \ref{section:super_point} \& \ref{section:segment_graph}
& 43.4 & 30.6 & 289 & 214 & 31.9 & 26.5 & 68.7\\
& 6. Ours  w/o \ref{section:super_point} \& \ref{section:instance_refinement}
& 41.7 & 29.3 & 282 & 208 & 30.4 & 27.1 & 68.7\\
& 7. Ours  w/o \ref{section:segment_graph} \& \ref{section:instance_refinement}
& 47.3 & 33.4 & 302 & 225 & 35.0 & 28.4 & \textbf{82.5} \\
\cdashline{2-9}
& 8. Ours w/ GT Mask
& 68.8 & 47.9 & 387 & 223 & 56.1 & 38.7 & 87.4 \\
\hline
\multirow{8}{*}{\rotatebox[origin=c]{90}{\scriptsize SLAM}} 
& 1. Ours
& \textbf{41.4} & \textbf{30.7} & \textbf{304} & \textbf{216} & \textbf{23.7} & \textbf{17.4} & \textbf{74.3} \\
& 2. Ours  w/o \ref{section:super_point}
& 33.0 & 24.2 & 229 & 182 & 18.5 & 14.8 & 63.7 \\
& 3. Ours  w/o \ref{section:segment_graph}
& 39.1 & 28.9 & 283 & 209 & 22.6 & 16.9 & \textbf{74.3}\\
& 4. Ours  w/o \ref{section:instance_refinement}
& 38.5 & 27.5 & 278 & 201 & 21.7 & 16.2 & \textbf{74.3}\\
& 5. Ours  w/o \ref{section:super_point} \& \ref{section:segment_graph}
& 34.9 & 22.2 & 237 & 176 & 17.6 & 14.6 & 63.7 \\
& 6. Ours  w/o \ref{section:super_point} \& \ref{section:instance_refinement}
& 30.7 & 21.5 & 218 & 172 & 15.9 & 13.7 & 63.7 \\
& 7. Ours  w/o \ref{section:segment_graph} \& \ref{section:instance_refinement}
& 37.4 & 28.3 & 273 & 205 & 19.8 & 16.1 & \textbf{74.3}\\
\cdashline{2-9}
& 8. Ours w/ GT Mask
& 49.6 & 37.2 & 318 & 253 & 38.2 & 28.4 & 79.7 \\
\hline
\end{tabular}}
\end{subtable}

\end{center}
\vspace{-10pt}
\end{table*}

%
The pipeline is implemented in C++ and runs on an Intel-12700K CPU.
The RGB panoptic segmentation runs on an Nvidia RTX3090. We run \cite{Semantic:voxbloxplusplus, ModifiedVoxblox} on same machine, while \cite{INS_CONV} is run on an E5-2680 CPU and Titan Xp GPU due to library dependency issue.
For \textit{all} sequences in SceneNN and ScanNet v2, we use the same hyper-parameters: 
$\theta_{merge}=3$ (Sec.~\ref{section:super_point}),
$K_C=15.0, \theta=0.5$ (Sec.~\ref{section:segment_graph}),
Parameters 
$\theta^C_{O}$, $\theta^C_{D}$, $\theta^C_{L}$ (Sec.~\ref{section:instance_refinement}) are tuned on ScanNet for 18 semantic categories.
The Voxel-TSDF voxel size is $1$cm.

\subsection{Experiments on SceneNN} 
\label{sssec:scene_nn_experiment}

As \cite{VoxbloxDiffusion, MultiviewFusion, IncrementalBBox} only report mAP with IoU thresholded at 0.5 and their code is not available, we follow their evaluation setting on 10 indoor sequences from SceneNN, considering 9 object categories (bed, chair, sofa, table, books, fridge, TV, toilet, and bag), and report $\text{mAP}_{50}$. 
For fair comparison, the CAD post-processing from Han et al. \cite{ModifiedVoxblox} is not used since it needs a CAD object database and replaces the predicted instances with models from the database.

The results are shown in Table \ref{table:scenenn_10}.
In the upper section, we evaluate methods using GT camera poses as input. 
The proposed method stands out, achieving the state-of-the-art average mAP50.
When we switch to SLAM poses as input (lower section), there is a notable decline in the accuracy across all methods, with the average mAP50 reducing by ${\approx}20$.
Even under these conditions, our algorithm remains the most accurate one. Notably, the gap between our method and the second best (INS-CONV) widens, underscoring the robustness and accuracy of the proposed approach.

\par
To avoid being restricted to the 10 scenes used in prior work, we ran the methods on the entire validation set comprising 20 sequences, considering all annotated objects of 18 categories. 
The per-class metrics are averaged over each instance and then averaged over categories for class-averaged metrics.
The results shown in Table \ref{table:scenenn_10_val} highlight that, when the GT camera poses serve as the input, our algorithm slightly trails behind INS-CONV in metrics like $\text{mAP}$, $N_\text{TP}$, and $\text{PQ}$. 
However, the dynamics change as we move to the lower part of the table, which reports results with SLAM poses as input.
In this practical scenario, our algorithm asserts its superiority, notably outperforming all methods in terms of $\text{mAP}$, $N_\text{TP}$, and $\text{PQ}$ and IoU$_{L_S}$ scores. 
This further underscores the criticality of using estimated poses over ground truth ones.

\begin{table*}
\setlength\extrarowheight{0.5pt}
\centering
\caption{Results on the validation set (200 scenes) of the ScanNet v2 dataset \cite{ScanNet} using the GT (top) and SLAM-estimated~\cite{campos2021orb} camera poses. 
The reported metrics are the class-averaged $\text{mAP}$, panoptic quality (\text{PQ}), and class-summed $N_\text{TP}$.
Also, we show our results with the ground truth instance segmentation mask as an accuracy upper bound.
}
%
\label{table:scannet_1513}
\vspace{-2mm}
\begin{center}
\begin{tabular}{c | l | c c | c c | c c }
\hline
& Method  &  $\text{mAP}_{50} $  &  $\text{mAP}_{75} $ ($\uparrow$) & $N_{\text{TP}_{50}}$ & $N_{\text{TP}_{75}}$ ($\uparrow$) & $PQ_{50}$ & $PQ_{75}$ ($\uparrow$)  \\  
\hline
\multirow{5}{*}{\rotatebox[origin=c]{90}{\scriptsize GT trajectory}} 
& Voxblox++ \cite{Semantic:voxbloxplusplus} 
& 47.2 & 23.9 & 2493 & 1408 & 38.6  & 21.7  \\

& Han et al. \cite{ModifiedVoxblox}
& 49.7 & 25.8 & 2586 & 1476 & 40.3  & 24.4  \\

& INS-CONV \cite{INS_CONV}
& \textbf{57.4} & \textbf{30.2} & \textbf{2917} & \textbf{1736} & \textbf{48.2}  & 28.5 \\

& Ours
& 55.5 & 29.8 & 2862 & 1672 & 46.8  &\textbf{29.1} \\ \cdashline{2-8}
& Ours w/ GT Mask
& 71.6 & 44.7 & 3539 & 1362 & 57.8  &41.5 \\
\hline
\multirow{5}{*}{\rotatebox[origin=c]{90}{\scriptsize SLAM}} 
& Voxblox++ \cite{Semantic:voxbloxplusplus} 
& 21.9 & 14.1 & 1389 & 692 & 23.2 & 13.4 \\
& Han et al. \cite{ModifiedVoxblox}
& 25.7 & 15.4 & 1479 & 754 & 23.9 & 15.8 \\
& INS-CONV \cite{INS_CONV}
& 29.4 & 17.6 & 1764 & 803 & 25.4 & 16.9 \\
& Ours
& \textbf{32.9} & \textbf{20.8} & \textbf{1902} & \textbf{932} & \textbf{31.2} & \textbf{20.5} \\
\cdashline{2-8}
& Ours w/ GT Mask
& 48.6 & 26.1 & 2512 & 1520 & 39.4 & 26.7 \\
\hline
\end{tabular}

\end{center}
\vspace*{-4mm}
\end{table*}

\subsection{Experiment on ScanNet v2}
We run experiments on the validation set of ScanNet v2 \cite{ScanNet} consisting of 200 sequences.
The results in Table \ref{table:scannet_1513} exhibit a similar trend as the ones on SceneNN. 
Running methods on the GT poses (upper part) leads to accurate results, with the proposed method being the second best in most metrics, trailing slightly behind INS-CONV.
Using SLAM poses, as would be done in practice, significantly reduces the accuracy of all methods and changes the ranking.
In this scenario, the proposed method outperforms the recent INS-CONV in \textit{all} metrics, showcasing the importance of using estimated camera trajectories.

We also ran the proposed algorithm using the GT 2D instance segmentation masks to provide an upper bound on the accuracy when having perfect 2D panoptic segmentation.
The accuracy significantly improves in both the GT and SLAM camera pose cases, showcasing that the proposed algorithm can be straightforwardly and substantially enhanced with the development of new panoptic networks. 
Let us note that we did not run other methods with GT masks, as, with this experiment, we only wanted to highlight the dependency on the panoptic segmentation network. 

%

\subsection{Ablation Study}

Table~\ref{table:scenenn_10_ablation} displays the ablation study with components removed from the proposed method. 
Label "w/o \ref{section:super_point}" means we use the super-point segmentation module from \cite{ModifiedVoxblox} instead of the proposed one; 
"w/o III-C and w/o III-D" means no semantic regularization and no instance refinement, respectively; 
"w/ GT Mask" means we use ground truth panoptic masks in \ref{section:2d_segmentation} by rendering from GT meshes instead of using predictions from \cite{mask2former}.
We show results both in the GT pose and SLAM cases.
While the two cases have a significant performance gap, ranking the components by importance seems similar.
The conclusions are as follows:
\begin{enumerate}[leftmargin=12pt]
    \item \textbf{Method 1 compared to 2, 3 to 5 and 4 to 6.} The results show that making the super-points semantically consistent (Section~\ref{section:super_point}) is crucial to improve the accuracy of semantic-instance segmentation. 
    Besides other error metrics, this is explicitly shown by the super-point IoU$_{L_S}$, which significantly reduces without this method, qualitative results are shown in Fig. \ref{fig:ablation_super-point}.
    \item \textbf{Method 1 compared to 3, 1 to 4 and 1 to 7.} The results show that the methods from both Sections~\ref{section:segment_graph} and \ref{section:instance_refinement} play significant roles in improving the accuracy of semantic-instance segmentation, qualitative results are shown in Fig. \ref{fig:ablation_graph_based_seg}.  
    \item \textbf{Method 1 compared to 8.} The results shows that the 2D panoptic predictions from \cite{mask2former} is one of the main factors that limit our semantic segmentation accuracy. Our method improves significantly with more accurate 2D panoptic predictions as input.
\end{enumerate}
\subsection{Run-time and Memory Analysis}
Table. \ref{table:runtime} shows the run-time and memory analysis of each component in the pipeline.
The 2D panoptic-geometric segmentation and fusion in Section \ref{section:2d_segmentation} takes the majority of the time and memory usage.
It is important to note that this largely depends on the employed panoptic segmentation network.
Thus, changing the network substantially changes the required time and memory. 
With limited computational resources, a lightweight panoptic segmentation (e.g., \cite{LPSNet}) and lower voxel resolution can be used.
%
%

Table~\ref{table:framerate} compares the average framerates with \cite{ModifiedVoxblox, Semantic:voxbloxplusplus, INS_CONV}.
Similar to \cite{Semantic:voxbloxplusplus}, \ref{section:2d_segmentation} is put into a separate thread and runs parallel to the main mapping thread to speed up.
The proposed method leads to similar framerates to other methods, all running in real-time. 
We believe it can still be accelerated by further code optimization. Note that semantic regularization and instance refinement run only once at the end of mapping for mesh generation. 




\begin{table}[H]
\centering
\caption{Run-time and memory analysis on the 10 sequences of SceneNN dataset \cite{SceneNN} evaluated in Table \ref{table:scenenn_10}.}
\label{table:run_expense}
\vspace{-2mm}
\begin{center}

\begin{subtable}{1.0\linewidth}\centering
\caption{Execution Times of Components}\label{table:runtime}
\vspace{-1mm}
\scriptsize
\begin{tabular}{l | c | c}
\hline
Component  & Run-Time ($ms$) & Memory Usage (MB)\\  
\hline
Sec.~\ref{section:2d_segmentation} &
216.0 (each frame) & \phantom{11}5014\phantom{} + 4359 (GPU) \\
Sec.~\ref{section:super_point} & \phantom{1}70.3 (each frame) & \phantom{1111}68.1\phantom{111111} \\
Sec.~\ref{section:segment_graph} (graph update) & 127.2 (each frame) & \phantom{1111}33.8\phantom{111111} \\
Sec.~\ref{section:segment_graph} (semantic reg.) & 324.0 (only once)\phantom{1} & \phantom{11111}0.3\phantom{111111} \\
Sec.~\ref{section:instance_refinement} (instance ref.) & \phantom{11}9.4 (only once)\phantom{1} & \phantom{11111}0.1\phantom{111111} \\
Label-TSDF Map & -- & [1530.3, 2072.0] \\ 
\hline
\end{tabular}
\end{subtable}

\begin{subtable}{1.0\linewidth}\centering
\vspace{2mm}
\caption{Average Frame-Rate Comparison}\label{table:framerate}
\scriptsize

\begin{tabular}{l | c c c c }
\hline
Method  & Voxblox++ \cite{Semantic:voxbloxplusplus} & Han \cite{ModifiedVoxblox} & INS-CONV \cite{INS_CONV} & Ours \\  
\hline
Framerate (Hz) &  3.82 & 3.66 & 5.46 & 3.76 \\
\hline
\end{tabular}
\end{subtable}
\newline

\vspace{-10pt}
\end{center}
\end{table}

%% file: root.bbl
\begin{thebibliography}{10}
\providecommand{\url}[1]{#1}
\csname url@rmstyle\endcsname
\providecommand{\newblock}{\relax}
\providecommand{\bibinfo}[2]{#2}
\providecommand\BIBentrySTDinterwordspacing{\spaceskip=0pt\relax}
\providecommand\BIBentryALTinterwordstretchfactor{4}
\providecommand\BIBentryALTinterwordspacing{\spaceskip=\fontdimen2\font plus
\BIBentryALTinterwordstretchfactor\fontdimen3\font minus
  \fontdimen4\font\relax}
\providecommand\BIBforeignlanguage[2]{{%
\expandafter\ifx\csname l@#1\endcsname\relax
\typeout{** WARNING: IEEEtran.bst: No hyphenation pattern has been}%
\typeout{** loaded for the language `#1'. Using the pattern for}%
\typeout{** the default language instead.}%
\else
\language=\csname l@#1\endcsname
\fi
#2}}

\bibitem{Semantic:voxbloxplusplus}
M.~{Grinvald}, F.~{Furrer}, T.~{Novkovic}, J.~J. {Chung}, C.~{Cadena},
  R.~{Siegwart}, and J.~{Nieto}, ``{Volumetric Instance-Aware Semantic Mapping
  and 3D Object Discovery},'' \emph{IEEE Robotics and Automation Letters
  (RA-L)}, July 2019.

\bibitem{ModifiedVoxblox}
M.~Han, Z.~Zhang, Z.~Jiao, X.~Xie, Y.~Zhu, S.-C. Zhu, and H.~Liu,
  ``Reconstructing interactive 3d scenes by panoptic mapping and cad model
  alignments,'' \emph{International Conference on Robotics and Automation},
  2021.

\bibitem{INS_CONV}
L.~Liu, T.~Zheng, Y.~Lin, K.~Ni, and L.~Fang, ``Ins-conv: Incremental sparse
  convolution for online 3d segmentation,'' \emph{IEEE/CVF Conference on
  Computer Vision and Pattern Recognition (CVPR)}, 2022.

\bibitem{Han20CVPR}
L.~Han, T.~Zheng, L.~Xu, and L.~Fang, ``{OccuSeg: Occupancy-aware 3D Instance
  Segmentation},'' \emph{IEEE/CVF Conference on Computer Vision and Pattern
  Recognition (CVPR)}, 2020.

\bibitem{Wang19CVPR}
X.~Wang, S.~Liu, X.~Shen, C.~Shen, and J.~Jia, ``{Associatively Segmenting
  Instances and Semantics in Point Clouds},'' \emph{IEEE/CVF Conference on
  Computer Vision and Pattern Recognition (CVPR)}, 2019.

\bibitem{Elich19GCPR}
C.~Elich, F.~Engelmann, T.~Kontogianni, and B.~Leibe, ``{3D Bird’s-eye-view
  Instance Segmentation},'' \emph{The German Conference on Pattern Recognition
  (GCPR)}, 2019.

\bibitem{Lahoud19ICCV}
J.~Lahoud, B.~Ghanem, M.~Pollefeys, and M.~R. Oswald, ``{3D Instance
  Segmentation via Multi-Task Metric Learning},'' \emph{International
  Conference on Computer Vision (ICCV)}, 2019.

\bibitem{miao2021tianjiport}
Y.~Miao, C.~Li, Z.~Li, Y.~Yang, and X.~Yu, ``A novel algorithm of ship
  structure modeling and target identification based on point cloud for
  automation in bulk cargo terminals,'' \emph{Measurement and Control}, 2021.

\bibitem{Hou19CVPR}
J.~Hou, A.~Dai, and M.~Nie{\ss}ner, ``{3D-SIS: 3D Semantic Instance
  Segmentation of RGB-D Scans},'' \emph{IEEE/CVF Conference on Computer Vision
  and Pattern Recognition (CVPR)}, 2019.

\bibitem{Yang19NIPS}
B.~Yang, J.~Wang, R.~Clark, Q.~Hu, S.~Wang, A.~Markham, and N.~Trigoni,
  ``{Learning Object Bounding Boxes for 3D Instance Segmentation on Point
  Clouds},'' \emph{Neural Information Processing Systems (NeurIPS)}, 2019.

\bibitem{Chen21ICCV}
S.~Chen, J.~Fang, Q.~Zhang, W.~Liu, and X.~Wang, ``{Hierarchical Aggregation
  for 3D Instance Segmentation},'' \emph{International Conference on Computer
  Vision (ICCV)}, 2021.

\bibitem{Engelmann20CVPR}
F.~Engelmann, M.~Bokeloh, A.~Fathi, B.~Leibe, and M.~Nie{\ss}ner, ``{3D-MPA:
  Multi-Proposal Aggregation for 3D Semantic Instance Segmentation},''
  \emph{IEEE/CVF Conference on Computer Vision and Pattern Recognition (CVPR)},
  2020.

\bibitem{Jiang20CVPR}
L.~Jiang, H.~Zhao, S.~Shi, S.~Liu, C.-W. Fu, and J.~Jia, ``{PointGroup:
  Dual-Set Point Grouping for 3D Instance Segmentation},'' \emph{IEEE/CVF
  Conference on Computer Vision and Pattern Recognition (CVPR)}, 2020.

\bibitem{Vu22CVPR}
T.~Vu, K.~Kim, T.~M. Luu, X.~T. Nguyen, and C.~D. Yoo, ``{SoftGroup for 3D
  Instance Segmentation on Point Clouds},'' \emph{IEEE/CVF Conference on
  Computer Vision and Pattern Recognition (CVPR)}, 2022.

\bibitem{schult2022mask3d}
J.~Schult, F.~Engelmann, A.~Hermans, O.~Litany, S.~Tang, and B.~Leibe, ``Mask3d
  for 3d semantic instance segmentation,'' \emph{arXiv preprint
  arXiv:2210.03105}, 2022.

\bibitem{sun2022superpoint}
J.~Sun, C.~Qing, J.~Tan, and X.~Xu, ``Superpoint transformer for 3d scene
  instance segmentation,'' \emph{arXiv preprint arXiv:2211.15766}, 2022.

\bibitem{PanopticFusion}
G.~Narita, T.~Seno, T.~Ishikawa, and Y.~Kaji, ``Panopticfusion: Online
  volumetric semantic mapping at the level of stuff and things,''
  \emph{IEEE/RSJ International Conference on Intelligent Robots and Systems
  (IROS)}, 2019.

\bibitem{panoptic_segmentation}
A.~Kirillov, K.~He, R.~Girshick, C.~Rother, and P.~Dollar, ``Panoptic
  segmentation,'' \emph{IEEE/CVF Conference on Computer Vision and Pattern
  Recognition (CVPR)}, June 2019.

\bibitem{SceneGraphFusion}
S.-C. Wu, J.~Wald, K.~Tateno, N.~Navab, and F.~Tombari, ``Scenegraphfusion:
  Incremental 3d scene graph prediction from rgb-d sequences,'' \emph{IEEE/CVF
  Conference on Computer Vision and Pattern Recognition (CVPR)}, June 2022.

\bibitem{GeometricSeg}
F.~Furrer, T.~Novkovic, M.~Fehr, A.~Gawel, M.~Grinvald, T.~Sattler,
  R.~Siegwart, and J.~Nieto, ``Incremental object database: Building 3d models
  from multiple partial observations,'' \emph{IEEE/RSJ International Conference
  on Intelligent Robots and Systems (IROS)}, 2018.

\bibitem{mask_rcnn}
K.~He, G.~Gkioxari, P.~Dollar, and R.~Girshick, ``Mask r-cnn,''
  \emph{International Conference on Computer Vision (ICCV)}, Oct 2017.

\bibitem{VoxbloxDiffusion}
R.~Mascaro, L.~Teixeira, and M.~Chli, ``Volumetric instance-level semantic
  mapping via multi-view 2d-to-3d label diffusion,'' \emph{IEEE Robotics and
  Automation Letters (RA-L)}, 2022.

\bibitem{campos2021orb}
C.~Campos, R.~Elvira, J.~J.~G. Rodr{\'\i}guez, J.~M. Montiel, and J.~D.
  Tard{\'o}s, ``Orb-slam3: An accurate open-source library for visual,
  visual--inertial, and multimap slam,'' \emph{IEEE Transactions on Robotics},
  2021.

\bibitem{mask2former}
B.~Cheng, I.~Misra, A.~G. Schwing, A.~Kirillov, and R.~Girdhar,
  ``Masked-attention mask transformer for universal image segmentation,''
  \emph{IEEE/CVF Conference on Computer Vision and Pattern Recognition (CVPR)},
  2022.

\bibitem{GraphCut}
Y.~Boykov, O.~Veksler, and R.~Zabih, ``Fast approximate energy minimization via
  graph cuts,'' \emph{IEEE Transactions on Pattern Analysis and Machine
  Intelligence}, 2001.

\bibitem{SceneNN}
B.-S. Hua, Q.-H. Pham, D.~T. Nguyen, M.-K. Tran, L.-F. Yu, and S.-K. Yeung,
  ``Scenenn: A scene meshes dataset with annotations,'' \emph{International
  Conference on 3D Vision}, 2016.

\bibitem{MultiviewFusion}
L.~Wang, R.~Li, J.~Sun, X.~Liu, L.~Zhao, H.~S. Seah, C.~K. Quah, and
  B.~Tandianus, ``Multi-view fusion-based 3d object detection for robot indoor
  scene perception,'' \emph{Sensors}, 2019.

\bibitem{IncrementalBBox}
W.~Li, J.~Gu, B.~Chen, and J.~Han, ``Incremental instance-oriented 3d semantic
  mapping via rgb-d cameras for unknown indoor scene,'' \emph{Discrete Dynamics
  in Nature and Society}, 2020.

\bibitem{ScanNet}
A.~Dai, A.~X. Chang, M.~Savva, M.~Halber, T.~Funkhouser, and M.~Nie{\ss}ner,
  ``Scannet: Richly-annotated 3d reconstructions of indoor scenes,''
  \emph{IEEE/CVF Conference on Computer Vision and Pattern Recognition (CVPR)},
  2017.

\bibitem{LPSNet}
W.~Hong, Q.~Guo, W.~Zhang, J.~Chen, and W.~Chu, ``Lpsnet: A lightweight
  solution for fast panoptic segmentation,'' \emph{IEEE/CVF Conference on
  Computer Vision and Pattern Recognition (CVPR)}, June 2021.

\end{thebibliography}
